%% file: main.tex
\documentclass[10pt,twocolumn,letterpaper]{article}

\usepackage[pagenumbers]{iccv} 


\usepackage{pifont}
\usepackage{tabularx}
\usepackage{tabularray}
\usepackage{float}

\newcolumntype{P}[1]{>{\centering\arraybackslash}p{#1}}

\usepackage{tikz}

\definecolor{iccvblue}{rgb}{0.21,0.49,0.74}
\usepackage[pagebackref,breaklinks,colorlinks,allcolors=iccvblue]{hyperref}

\def\paperID{6} 
\def\confName{ICCV}
\def\confYear{2025}

\title{AesCrop: Aesthetic-driven Cropping Guided by Composition}

\author{Yen-Hong Wong \quad Lai-Kuan Wong\\
Faculty of Computing and Informatics, Multimedia University\\
Persiaran Multimedia, Cyberjaya, Selangor, Malaysia.\\
{\tt\small 1211101392@student.mmu.edu.my \quad lkwong@mmu.edu.my}
}

\begin{document}
\maketitle
\input{sec/0_abstract}    
\input{sec/1_intro}
\input{sec/2_related_work}

\input{sec/3_method}

\input{sec/4_experiments}
\input{sec/5_limitations}
\input{sec/6_conclusion}
{
    \small
    \bibliographystyle{ieeenat_fullname}
    \bibliography{main}
}

\end{document}

%% file: sec/0_abstract.tex
\begin{abstract}
Aesthetic-driven image cropping is crucial for applications like view recommendation and thumbnail generation, where visual appeal significantly impacts user engagement. A key factor in visual appeal is composition—the deliberate arrangement of elements within an image. Some methods have successfully incorporated compositional knowledge through evaluation-based and regression-based paradigms. However, evaluation-based methods lack globality while regression-based methods lack diversity. Recently, hybrid approaches that integrate both paradigms have emerged, bridging the gap between these two to achieve better diversity and globality.  Notably, existing hybrid methods do not incorporate photographic composition guidance, a key attribute that defines photographic aesthetics. 
In this work, we introduce AesCrop, a composition-aware hybrid image-cropping model that integrates a VMamba image encoder, augmented with a novel Mamba Composition Attention Bias (MCAB) and a transformer decoder to perform end-to-end rank-based image cropping, generating multiple crops along with the corresponding quality scores. By explicitly encoding compositional cues into the attention mechanism, MCAB directs AesCrop to focus on the most compositionally salient regions. Extensive experiments demonstrate that AesCrop outperforms current state-of-the-art methods, delivering superior quantitative metrics and qualitatively more pleasing crops.
\end{abstract}

%% file: sec/1_intro.tex
\section{Introduction}

Image cropping is the process of removing unwanted outer areas from an image. To crop an image in a visually appealing way, a solid understanding of image composition principles is essential. These principles guide the arrangement of elements within an image, ensuring balance, focus, and aesthetic harmony, which in turn directs the viewer's attention to the intended subject and conveys the desired message or emotion. However, without sufficient knowledge of aesthetic and photographic principles, amateurs often produce undesirable and unappealing crops. 

\begin{figure}
    \centering
    \includegraphics[width=0.4\textwidth]{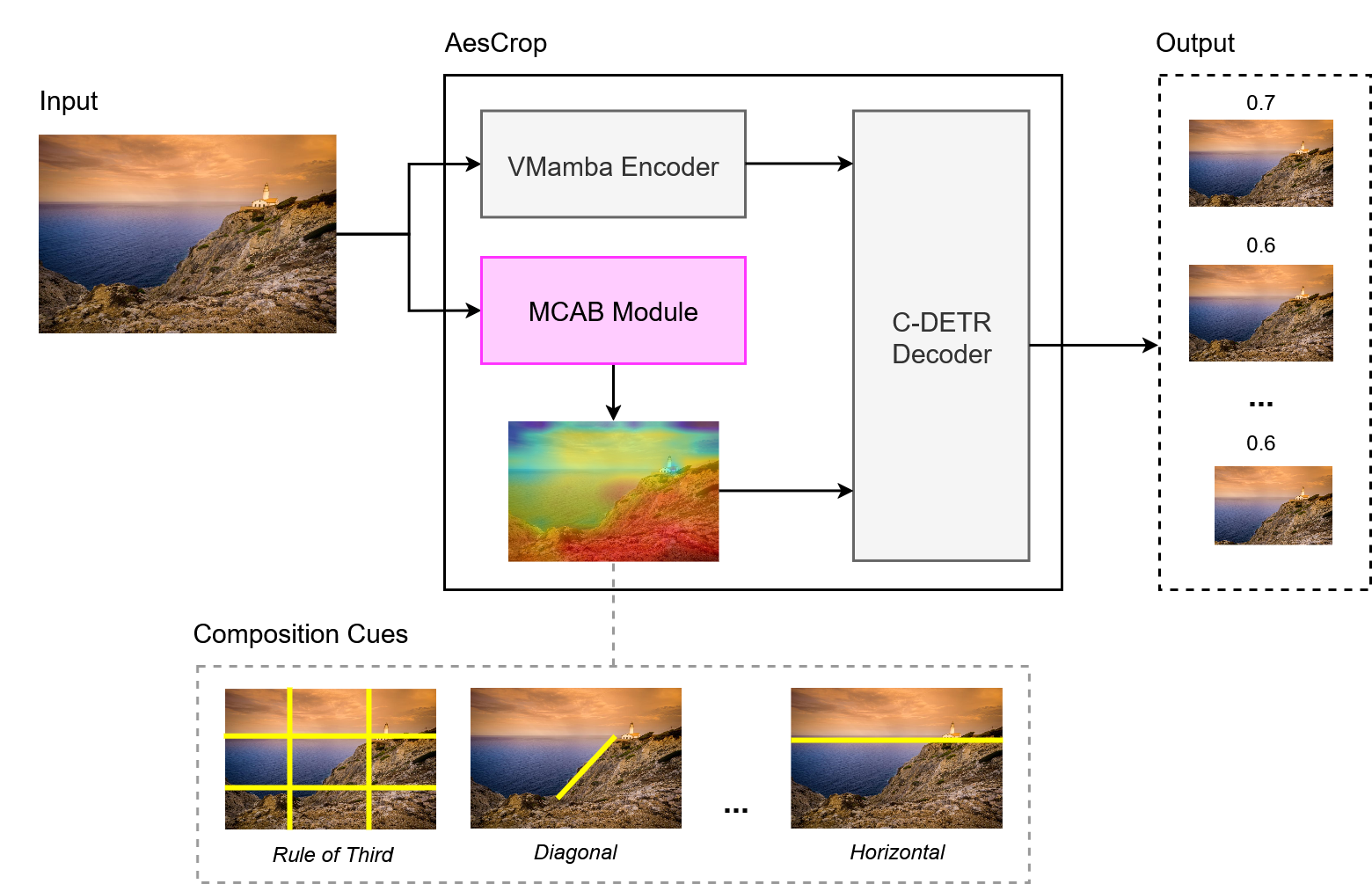}
    \caption{AesCrop is a composition-aware, hybrid image-cropping model that incorporates cues from multiple composition classes via the Mamba Composition Attention Bias (MCAB) and performs end-to-end rank-based image cropping.}
    \label{fig:teaser}
\end{figure}

To address this gap, researchers introduced automatic image cropping to make aesthetic-driven cropping more accessible. Early image cropping approaches can be categorized primarily into two paradigms: evaluation-based \cite{6618974, chen-acmmm-2017, wei2018good, zeng2019gridanchorbasedimage, YUAN2024104316, 9710744} and regression-based \cite{9578088, 8259308, pan2023beautyrarecontrastivecomposition, 10222223, hong2024learningsubjectawarecroppingoutpainting} methods. Evaluation-based approaches generate a fixed number of candidate crops using heuristic rules, evaluate their aesthetic quality, and rank them using a separate aesthetics evaluation model. Regression-based methods directly regress the best candidate crop. In particular, incorporation of explicit compositional knowledge into image-cropping methods has been proven effective in enhancing the aesthetic appeal of cropped images in both evaluation-based \cite{YUAN2024104316} and regression-based \cite{9578088,10222223} approaches. However, both paradigms have limitations: the former lacks globality as heuristic-based crop candidates might miss the optimal choice, while the latter lacks diversity since generating a single crop may be insufficient. Recently, hybrid approaches \cite{9878802, 10350920} that simultaneously regress and evaluate multiple candidate crops have been proposed to achieve better diversity and globality. These methods can generate pleasing crops but none consider explicit compositional guidance, which could potentially further enhance the aesthetics quality of cropped images. 

In this work, we introduce AesCrop, a novel, composition-aware hybrid cropping model that incorporates explicit compositional cues to guide cropping, as illustrated in \cref{fig:teaser}. Capitalizing on the strength of Mamba's 
encoder in its efficient long-range dependency modeling with linear computational complexity, we 
integrates a VMamba-based encoder \cite{liu2024vmambavisualstatespace} that generates rich embedding of the input image with a Conditional DETR-based transformer decoder \cite{DBLP:journals/corr/abs-2108-06152} that regresses and scores multiple candidate crops using learnable anchor embeddings. To incorporate explicit compositional knowledge, we propose the Mamba Composition Attention Bias (MCAB), a novel mechanism that extracts composition cues from a VMamba-based \cite{liu2024vmambavisualstatespace} composition classifier and injects them into the transformer decoder by modulating its attention weights. MCAB guides the decoder to focus on compositionally salient regions, yielding crops that adhere to established compositional principles. 

In summary, our main contributions are as follows:

\begin{itemize}
    \item We present AesCrop, a novel hybrid image cropping model that incorporates explicit compositional knowledge to simultaneously achieve globality, diversity, and compositional awareness. Specifically, Mamba Composition Attention Bias (MCAB), a novel mechanism that extracts compositional cues from a Mamba-based classifier and injects them into attention weights is introduced to effectively guide the model to focus on compositionally important regions.
    \item Comprehensive experiments and ablation studies validate our architectural design choices and demonstrate that AesCrop consistently outperforms state-of-the-art methods in generating aesthetically pleasing crops that adhere more faithfully to professional photographic composition rules.
\end{itemize}

%% file: sec/2_related_work.tex
\section{Related Work}

Existing approaches can be broadly categorized into three paradigms: evaluation-based, regression-based, and hybrid-based methods. 

\textbf{Evaluation-Based}. In this paradigm, the majority of works follow a two-step process: (1) generating a number of candidate crops using heuristic rules, and (2) ranking each candidate crop to determine the best crops. Within this framework, several representative approaches have pushed the state of the art by improving either candidate generation or crop scoring. Yan \etal \cite{6618974} proposed a method that uses regional features taking changes into account to generate candidate crops, then scores these crops with an SVM. VFN \cite{chen-acmmm-2017} assumes that web images have the best aesthetics, allowing the model to learn from large-scale pairwise ranking in an unsupervised manner. VPN \cite{wei2018good} learns to score image crops in real-time using a novel knowledge distillation framework. GAIC \cite{zeng2019gridanchorbasedimage} reduces the candidate search space using special properties of image cropping, then ranks the crops using a CNN that models the RoI and RoD. TransView \cite{9710744} utilizes a transformer-based architecture to model dependencies between visual elements inside and outside the bounding box for crop ranking. CLIPCropping \cite{YUAN2024104316} leverages CLIP to learn aesthetic and composition concepts from dedicated image-text pairs for crop ranking.

\textbf{Regression-Based}. In this paradigm, most approaches directly regress the coordinates of the optimal crop through a single-stage process. Within this paradigm, several methods have driven progress by refining regression strategies or by injecting domain knowledge into the prediction process. Guo \etal \cite{8259308} introduced a cascaded regression method that progressively refines crop predictions through iterative adjustments. Mars \cite{9156674} adapts a base model to diverse aspect ratio requirements by employing two meta-learners that dynamically generate model parameters based on target aspect ratios, enabling flexible and precise cropping across various formats. CACNet \cite{9578088} utilizes Class Activation Maps (CAMs) to extract compositional features through a dedicated composition branch, which are then explicitly integrated into the cropping branch to enhance output quality. Pan \etal \cite{pan2023beautyrarecontrastivecomposition} operates on the principle that crops with similar boundaries share compositional characteristics, proposing C2C to regularize these features for boundary prediction that generalizes across both common and rare patterns. GenCrop \cite{hong2024learningsubjectawarecroppingoutpainting} builds on the assumption that stock images exhibit optimal aesthetics, training its model on a large-scale image-crop dataset through a diffusion-based outpainting approach.

\textbf{Hybrid-Based}. The paradigm was first introduced by Jia et al. \cite{9878802}, who rethink image cropping through a hybrid method using Conditional DETR with a novel training strategy adapted from DETR. Their work introduced two label smoothing techniques to provide enhanced learning signals for crop ranking. Later, ClipCrop \cite{10350920} extended this approach by incorporating a CLIP image encoder and adding conditioned embeddings to the decoder's learnable anchors, enabling text- and image-conditional cropping. Notably, neither of these works in the hybrid paradigm has explicitly incorporated compositional knowledge.

%% file: sec/3_method.tex
\section{Methodology}

\begin{figure*}[htbp]
  \centering
  \includegraphics[width=0.9 \textwidth]{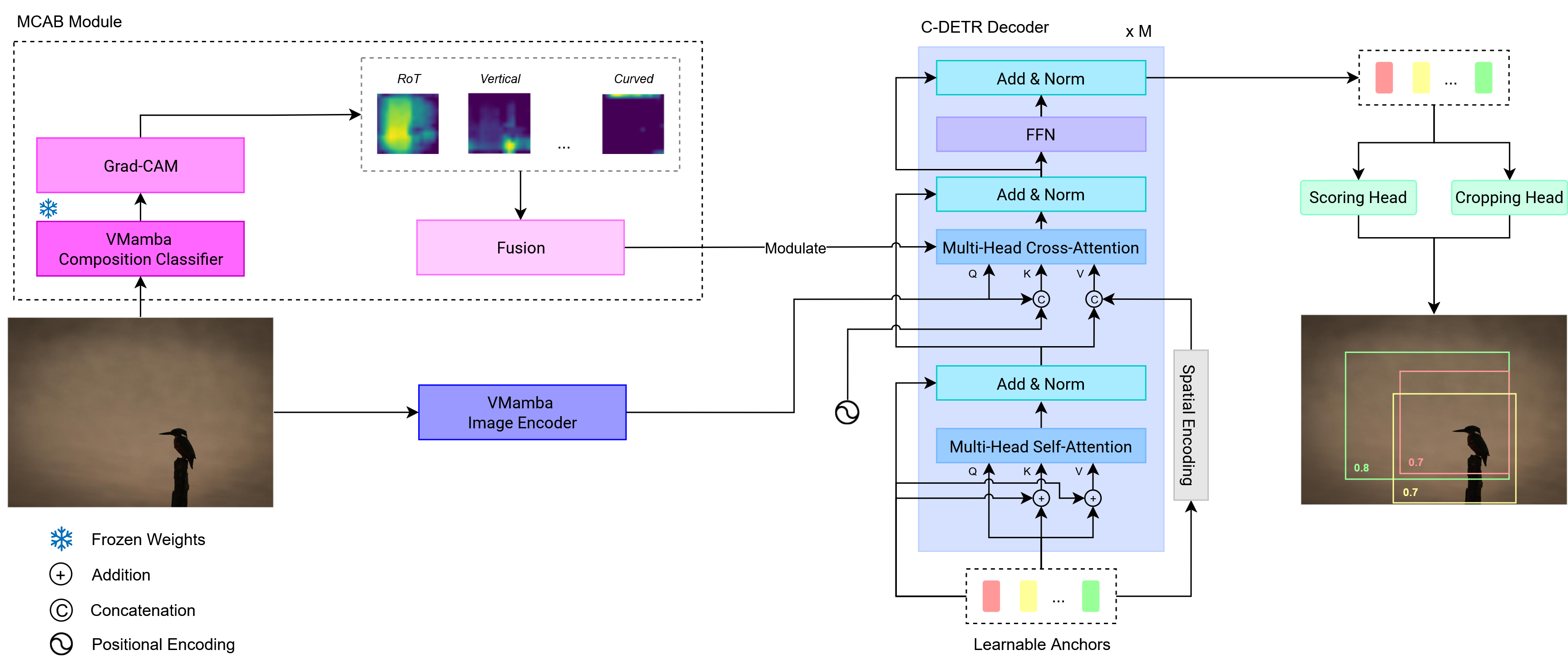}
  \vspace{-0.5em}
  \caption{\textbf{The architecture of AesCrop}. The input image is processed by dual-stream encoders: (1) a module that generates the Mamba Composition Attention Bias (MCAB) and (2) an encoder that produces image embeddings. A decoder then refines learnable queries via self-attention, aggregates those queries with the image embeddings through MCAB-modulated cross-attention, and processes the result through feed-forward networks. The final embeddings are passed to prediction heads that output crop boxes and quality scores.}
  \label{fig:aescrop_architecture}
\end{figure*}

To address the lack of compositional guidance in the hybrid-based paradigm, 
we propose AesCrop, a novel hybrid-based, composition-aware image cropping framework. 
AesCrop features dual-stream VMamba encoders \cite{liu2024vmambavisualstatespace} that simultaneously generate the novel Mamba Composition Attention Bias (MCAB) and image embeddings, along with a Conditional DETR-based decoder inspired by the design of Jia et al. \cite{9878802}. The complete architecture is illustrated in \cref{fig:aescrop_architecture}.

Our framework begins with a dual-stream encoding architecture designed to jointly capture both visual semantics and compositional structure. The visual processing stream utilizes a VMamba image encoder $\mathcal{F}_{\text{enc}}$ that transforms the input image $\mathbf{I}$ into comprehensive image embeddings $\mathbf{E}$, which capture essential visual features for crop selection:
\begin{equation}
    \mathbf{E} = \mathcal{F}_{\text{enc}}(\mathbf{I})
\end{equation}

Simultaneously, the compositional stream generates the composition attention bias, denoted by $\mathbf{B}$ through the $\mathcal{F}_{\text{MCAB}}$ module, encoding explicit compositional guidance:
\begin{equation}
    \mathbf{B} = \mathcal{F}_{\text{MCAB}}(\mathbf{I})
\end{equation}

These components, along with the learnable anchor queries $\mathbf{Q} = \{\mathbf{q}_i\}_{i=1}^N$, are processed by a Conditional DETR (C-DETR)  \cite{DBLP:journals/corr/abs-2108-06152} decoder $\mathcal{F}_{\text{dec}}$ with $M$ layers. Each decoder layer first applies self-attention to model global dependencies among the learnable anchors queries, establishing their contextual relationships. The layer then performs MCAB-modulated cross-attention, where the contextualized queries interact with the image embeddings $\mathbf{E}$ under the guidance of the compositional bias $\mathbf{B}$ generated by MCAB. This attention mechanism dynamically prioritizes compositionally salient regions by adjusting attention weights based on the composition priors from $\mathbf{B}$. Following these cross attention operations, the features undergo non-linear transformation through feed-forward networks (FFNs). Through $M$ decoder layers, the decoder progressively integrates visual features, spatial positions, and compositional knowledge into the final output embeddings $\mathbf{O}=\{\mathbf{o}_i\}_{i=1}^N$, where $\mathbf{o}_i$ represents the refined embedding for a candidate crop:
\begin{equation}
    \mathbf{O} = \mathcal{F}_{\text{dec}}(\mathbf{E}, \mathbf{B}, \mathbf{Q})
\end{equation}
The final processing stage employs two specialized prediction heads that operate on each output embedding $\mathbf{o}_i$. The crop prediction head $f_{\text{crop}}$ generates normalized bounding box coordinates $\hat{\mathbf{b}}_i \in [0,1]^4$ for each candidate crop, consisting of the predicted center coordinates ($\hat{c}_x^i$, $\hat{c}_y^i$) and dimensions ($\hat{w}^i$, $\hat{h}^i$). Simultaneously, the scoring head $f_{\text{score}}$ produces a normalized quality score $\hat{v}_i \in [0,1]$ for the corresponding candidate crop:
\begin{align}
    f_{\text{crop}}(\mathbf{o}_i) &= \hat{\mathbf{b}}_i = [\hat{c}_x^i,\;\hat{c}_y^i,\;\hat{w}^i,\;\hat{h}^i] \\
    f_{\text{score}}(\mathbf{o}_i) &= \hat{v}_i
\end{align}

\subsection{Mamba Composition Attention Bias}

\begin{figure}
    \centering
    \begin{tabular}{m{2.3cm} m{2.3cm} m{2.3cm}}
    \multicolumn{1}{c}{\small Center} & 
    \multicolumn{1}{c}{\small Curved} & 
    \multicolumn{1}{c}{\small Diagonal} \\

    \includegraphics[width=2.3cm]{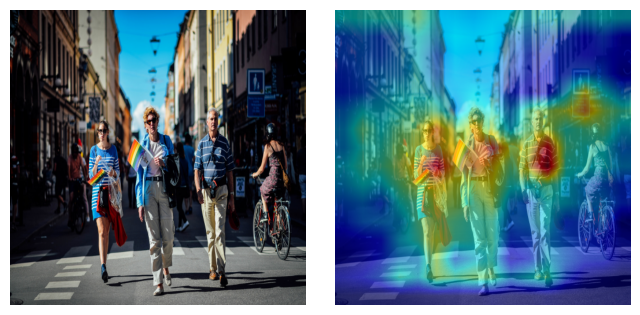} &
    \includegraphics[width=2.3cm]{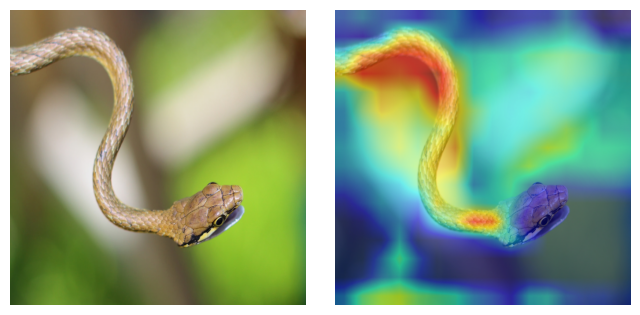} &
    \includegraphics[width=2.3cm]{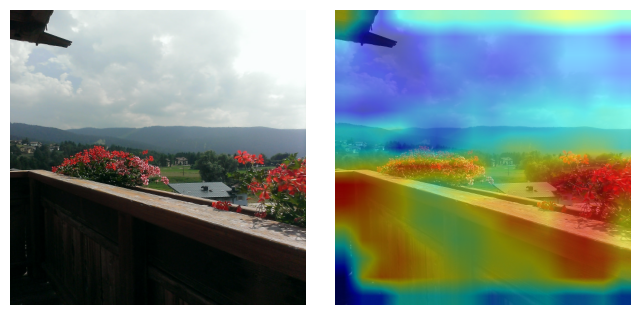} \\

    \multicolumn{1}{c}{\small Horizontal} & 
    \multicolumn{1}{c}{\small Pattern} & 
    \multicolumn{1}{c}{\small RoT} \\

    \includegraphics[width=2.3cm]{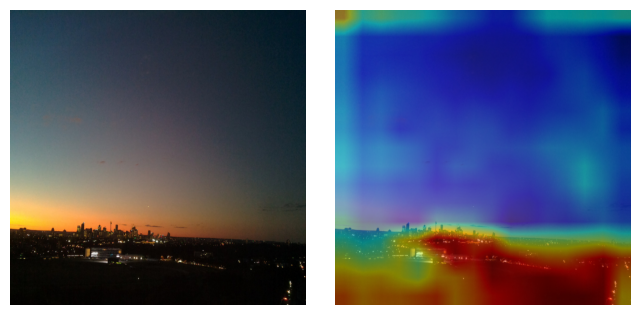} &
    \includegraphics[width=2.3cm]{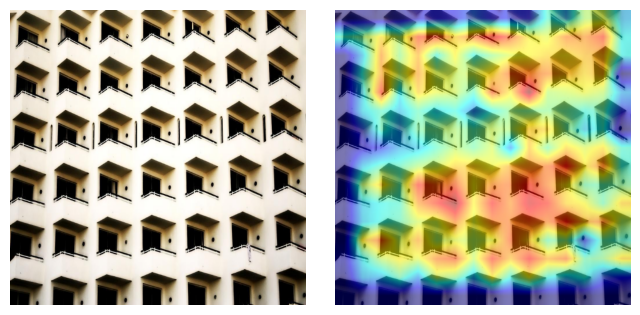} &
    \includegraphics[width=2.3cm]{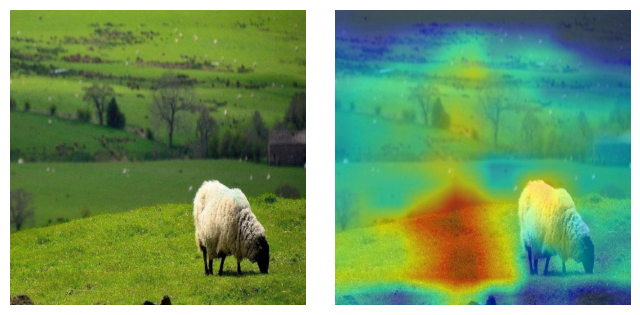} \\

    \multicolumn{1}{c}{\small Symmetric} & 
    \multicolumn{1}{c}{\small Triangle} & 
    \multicolumn{1}{c}{\small Vertical} \\

    \includegraphics[width=2.3cm]{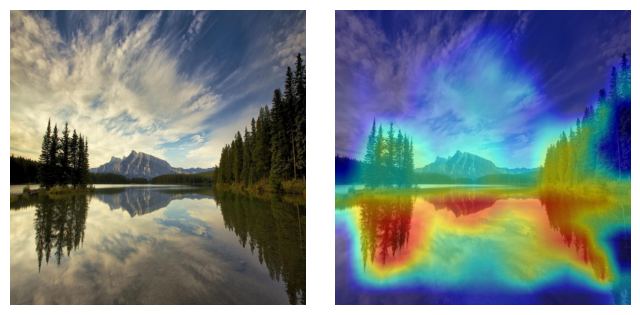} &
    \includegraphics[width=2.3cm]{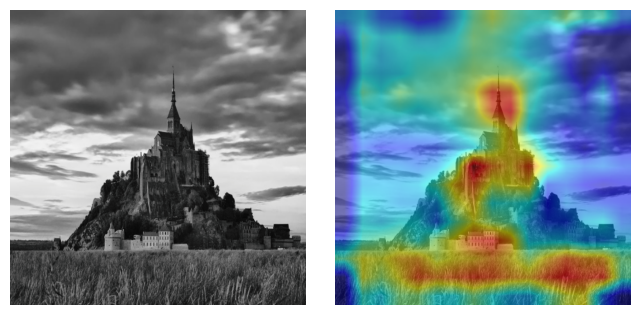} &
    \includegraphics[width=2.3cm]{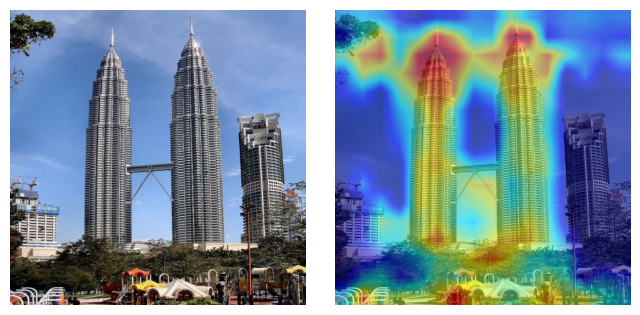} 
    \end{tabular}

    \vspace{-1em}
    \caption{Class activation maps of nine images representing distinct composition classes.} 
    \label{fig:composition_heatmap}
\end{figure}

The Mamba Composition Attention Bias (MCAB) module enables AesCrop to incorporate compositional knowledge by adaptively weighting compositionally salient regions. Specifically, the composition attention bias $\mathbf{B}$ is generated through a multi-step process via $\mathcal{F}_{\text{MCAB}}$. First, a frozen VMamba composition classifier pretrained on the KU-PCP dataset \cite{LEE201891}, which consists of nine image composition classes, is used to output a probability $p_i$ for each class $i$. For each predicted class $i$, Grad-CAM \cite{DBLP:journals/corr/SelvarajuDVCPB16} produces a corresponding Class Activation Map (CAM) $\mathbf{C}_i$ by computing gradients of the composition class score with respect to the feature maps. Each $\mathbf{C}_i$ is a heatmap that matches the dimension of the input image, with values normalized between [0, 1] from least important to most important, highlighting the regions most relevant to the composition class $i$. Examples of these CAMs are illustrated in \cref{fig:composition_heatmap}.

The CAMs are then aggregated through probability-weighted averaging in the fusion block to generate composition attention bias, $\mathbf{B}$:
\begin{equation}
\mathbf{B} = \sum_{i=1}^{N} p_i \cdot \mathbf{C}_i
\end{equation}
This operation produces $\mathbf{B}$, a unified global heatmap that captures the attentive regions corresponding to each composition class. Then, $\mathbf{B}$ is normalized to the range $[0, 1]$. The resulting $\mathbf{B}$ is downsampled and flattened to spatially align with the image encoder's output dimensions, yielding the final MCAB.

Traditionally, cross-attention between queries $\mathbf{Q}$, keys $\mathbf{K}$, and values $\mathbf{V}$ is computed as:
\begin{equation}
\text{softmax}\left(\frac{\mathbf{Q}\mathbf{K}^T}{\sqrt{d_k}}\right)\mathbf{V}
\end{equation}
where $\mathbf{Q}\mathbf{K}^T$ represents the attention weights, a matrix of shape $|\mathbf{Q}| \times |\mathbf{K}|$ whose element at the $i$-th row and $j$-th column indicates how much query $i$ attends to key $j$. In AesCrop, this determines how much learnable anchor $i$ focuses on image patch $j$.

To inject composition knowledge, we modify the cross-attention mechanism as follows:
\begin{equation}
\text{softmax}\left(\frac{\mathbf{Q}\mathbf{K}^T}{\sqrt{d_k}} + \log{\mathbf{B}}\right)\mathbf{V}
\end{equation}
Here, $\mathbf{B}$ undergoes log-scaling to amplify the weight of compositionally salient regions relative to non-salient ones, yielding $\log{\mathbf{B}}$. This log‐scaled term is then added to the original attention weights, biasing the model toward compositionally important regions by adjusting the attention of learnable anchor $i$ toward image patch $j$. This modification directs the decoder's focus to areas rich in compositional elements, significantly improving the model's ability to generate crop proposals that adhere to compositional principles.


\subsection{Hybrid Cropping Training Strategy}

Training a hybrid cropping network without fixed anchors or post-processing steps such as non-maximum suppression requires an effective matching strategy between predictions and ground truth crops. Following Jia \etal \cite{9878802}, we first isolate a set of high-quality ground truths, denoted $y_{\text{good}}$, by selecting those with mean opinion scores $s_i \ge 4$. Additionally, we pad $y_{\text{good}}$ with empty objects so as to match the cardinality of $\hat{y}$. Next, we employ the Hungarian Algorithm to establish a one-to-one correspondence $\hat\sigma$ between our predicted crops $\hat{y}$ and these high-quality annotations. This correspondence is chosen to minimize the aggregate matching loss:
\begin{equation}
    \hat{\sigma} = \underset{\sigma}{\arg\min}\;\sum_{i=1}^{N}\mathcal{L}_{\text{match}}\bigl(\hat{y}_i,\,y_{\text{good},\,\sigma(i)}\bigr)
\end{equation}
Here, the matching loss \(\mathcal{L}_{\text{match}}\) integrates three components:
\begin{equation}
\begin{split}
    \mathcal{L}_{\text{match}} (y_{\text{good}, i}, \hat{y}_{\sigma(i)}) &= \mathcal{L}_{\text{reg}}(\hat{b}_{i}, b_{\text{good}, \sigma(i)}) \\
    &\quad + \lambda_{\text{GIoU}} \mathcal{L}_{\text{GIoU}}(\hat{b}_{i}, b_{\text{good}, \sigma(i)}) \\
    &\quad + \lambda_{\text{focal}} \mathcal{L}_{\text{focal}}(\hat{v}_{i}, v_{\text{good}, \sigma(i)})
\end{split}
\end{equation}
where $\mathcal{L}_{\text{reg}}$, $\mathcal{L}_{\text{GIoU}}$, and $\mathcal{L}_{\text{focal}}$ are the L1 regression loss, the Generalized IoU loss, and the focal loss, respectively, with $\lambda_{\text{GIoU}}$ and $\lambda_{\text{focal}}$ as weighting hyperparameters.

For predicted crops matched to non‐empty, high‐quality ground truths, we supervise both their bounding boxes and quality scores using those matched annotations. However, the majority of predictions will be matched to an empty object. To provide additional learning signals for these predictions, any unmatched prediction that nevertheless overlaps substantially (IoU $\geq$ 0.85) with a ground truth crop is supervised by assigning it a soft label estimated from the quality score of its high‑IoU neighbor, thus exploiting local redundancy. Finally, for all remaining predictions that neither match a high‑quality crop nor exhibit substantial overlap, we set their quality scores to zero, providing explicit negative supervision for low‑quality crops.

%% file: sec/4_experiments.tex
\section{Experiments}

\subsection{Datasets}

We use GAIC \cite{zeng2019gridanchorbasedimage} dataset as our cropping dataset, as it is the only sufficiently densely annotated dataset suitable for training hybrid-based cropping models \cite{9878802}. Specifically, we adopt GAICv2, which offers a larger number of images compared to its predecessor, GAICv1. Additionally, we use KU-PCP \cite{LEE201891} as the composition dataset to train our VMamba composition classifier.

\textbf{GAICv2} \cite{zeng2019gridanchorbasedimage} contains 3,336 images, split into 2,636 training samples, 200 validation samples, and 500 test samples. Each image includes up to 90 candidate crops annotated with mean opinion scores (MOS) ranging from 1 to 5. The dataset demonstrates strong annotator agreement, with 94.25\% of crops having a score standard deviation below 1.

\textbf{KU-PCP} \cite{LEE201891} consists of 4,244 images, divided into 3,169 training samples and 1,075 testing samples. Each image is annotated with one or more of the nine composition classes: \textit{rule of thirds, center, horizontal, vertical, symmetric, diagonal, curved, triangle}, and \textit{pattern}.

\subsection{Evaluation Metrics}

We evaluate our method using the Return $K$ of Top $N$ accuracy metric, $Acc_{K/N}$ \cite{zeng2019gridanchorbasedimage}, adapted for hybrid-based approaches \cite{9878802}. The metric is formally defined as:

\begin{equation}
Acc_{K/N} = \frac{1}{TK} \sum_{i=1}^{T} \sum_{j=1}^{K} \mathbb{I} \left( \max_{g \in S_i(N)} \text{IoU}\left(c_{i,j}, g\right) \geq \epsilon \right)
\end{equation}

\noindent where $\mathbb{I}(\cdot)$ is an indicator function that returns 1 when the condition is true and 0 otherwise, $\epsilon$ is the IoU threshold for crop matching, $T$ is the total number of test examples, $c_{i,j}$ represents the top $j$-th predicted crop for the $i$-th example, and $S_i(N)$ denotes the set of top $N$ ground truth crops for the $i$-th example.

We additionally employ the averaged metric $\overline{Acc}_{N}$, computed as:

\begin{equation}
    \overline{Acc}_{N} = \frac{1}{|S_K|} \sum_{K \in S_K} Acc_{K/N}
\end{equation}

\noindent where $S_K$ represents the set of $K$ values under consideration, and $|S_K|$ denotes its cardinality.

\subsection{Implementation Details}

\hspace*{1.2em}\textbf{Dataset Configuration}: Both cropping and composition datasets undergo similar preprocessing pipelines, with cropping datasets requiring additional steps. All images are first resized to a fixed $512\times512$ resolution to preserve high-quality details for accurate predictions. During training, we apply multiple data augmentation techniques to enhance model robustness, including random variations in saturation, hue, and contrast. Following augmentation, we normalize all images using ImageNet's mean and standard deviation values to ensure training stability. For cropping datasets, we perform additional processing by normalizing ground truth coordinates to the [0,1] range and converting them to the standardized $c_x, c_y, w, h$ format.

\textbf{Model Pretraining}: AesCrop employs a two-stage pretraining approach before final training. First, we pretrain the composition classifier on KU-PCP \cite{LEE201891} and keep its weights frozen during main training to ensure stable generation of MCAB. Second, we pretrain the remaining components on COCO-minitrain \cite{HoughNet}, a carefully selected 25,000 image subset of the COCO object detection dataset. This addresses the limited sample size in existing cropping datasets while avoiding the computational burden of the full 100,000+ image COCO training set. The COCO-minitrain subset maintains the statistical properties of the complete dataset while offering efficient training. After both pretraining stages complete, we initialize AesCrop's corresponding components with the learned weights.

\textbf{Model Configuration}:  AesCrop predicts $N = 90$ crops per image, aligning with the maximum number of annotations available in the GAICv2 \cite{zeng2019gridanchorbasedimage} dataset. The architecture employs VMamba-T variant for its compactness and efficiency and $M=6$ decoder layers, with loss weights set to $\lambda_{\text{focal}} = \lambda_{\text{GIoU}} = 0.4$. The model is optimized using AdamW, which adapts learning rates dynamically for stable training. All experiments run on an NVIDIA RTX 4090 GPU with a batch size of 16. The initial learning rate of $10^{-4}$ is maintained for 40 epochs, then divided by ten for the final 10 epochs.

\subsection{Comparison with Existing Works}

\begin{table*}
    \centering
    \caption{Quantitative Comparison. The best results are marked in \textbf{bold} and the second best results are marked with \underline{underline}.}
    \vspace{-1em}
    \label{tab:quant_comp}
    \begin{tabular}{p{3cm} P{1cm} P{1cm} P{1cm} P{1cm} P{1cm} P{1cm} P{1cm} P{1cm} P{1cm} P{1cm}}
    \hline
    \multicolumn{1}{c}{Model} & 
    $Acc_{1/5}$ & 
    $Acc_{2/5}$ & 
    $Acc_{3/5}$ & 
    $Acc_{4/5}$ & 
    $\overline{Acc}_5$ & 
    $Acc_{1/10}$ & 
    $Acc_{2/10}$ & 
    $Acc_{3/10}$ & 
    $Acc_{4/10}$ & 
    $\overline{Acc}_{10}$ \\
    \hline
    VFN \cite{chen-acmmm-2017} & 26.6 & 26.5 & 26.7 & 25.7 & 26.4 & 40.6 & 40.2 & 40.3 & 39.3 & 40.1\\
    VPN \cite{wei2018good} & 36.0 & - & - & - & - & 48.5 & - & - & - & -  \\
    VEN \cite{wei2018good}& 37.5 & 35.0 & 35.3 & 34.2 & 35.5 & 50.5 & 49.2 & 48.4 & 46.4 & 48.6 \\
    GAIC \cite{zeng2019gridanchorbasedimage}& 
    68.2 & 64.3 & 61.3 & 58.5 & 63.1 & 84.4 & 82.7 & 80.7 & \underline{78.7} & 81.6
    \\
    TransView \cite{9710744} & 69.0 & \underline{66.9} & 61.9 & 57.8 & 63.9 & 85.4 & \underline{84.1} & \underline{81.3} & 78.6 & 82.4 \\ 
    CLIPCropping \cite{YUAN2024104316}& 70.0 & 66.7 & \underline{63.0} & \underline{60.0} & \underline{64.9} & \underline{87.5} & 83.7 & 80.5 & 78.5 & \underline{82.5} \\
    Jia \etal \cite{9878802} & \underline{72.0} & - & - & - & - & 86.0 & - & - & - & - \\ 
    \hline
    AesCrop (Ours) & \textbf{79.4} & \textbf{75.8} & \textbf{71.5} & \textbf{68.3} & \textbf{73.7} & \textbf{92.2} & \textbf{90.4} & \textbf{87.1} & \textbf{84.8} & \textbf{88.6} \\ \hline
    \end{tabular}
\end{table*}

\hspace*{1.2em}\textbf{Quantitative Comparison}: For quantitative evaluation, we employ the $Acc_{K/N}$ and $\overline{Acc}_{N}$ metrics for $K \in \{1, 2, 3, 4\}$ and $N \in \{5, 10\}$ on the GAICv2 \cite{zeng2019gridanchorbasedimage} testing set with $\epsilon = 0.90$, similar to Jia \etal's \cite{9878802} experimental settings. The quantitative results for the existing works were obtained directly from their papers, and the comparisons are summarized in \cref{tab:quant_comp}. The results show that the proposed model, AesCrop outperforms state-of-the-arts methods across all evaluated metrics, with a substantial performance gain of 7.4\% in $Acc_{1/5}$ and 4.7\% in $Acc_{1/10}$. This impressive performance demonstrates the effectiveness of the MCAB module is guiding AesCrop to consider composition salient regions in generating aesthetically appealing crops. 

\begin{figure*}
    \centering
    \begin{tabular}{m{2cm} | m{2cm} m{2cm} m{2cm} | m{2cm} m{2cm} m{2cm}}
    \multicolumn{1}{c}{Input} & 
    \multicolumn{1}{c}{GT 1} & 
    \multicolumn{1}{c}{GT 2} & 
    \multicolumn{1}{c}{GT 3} & 
    \multicolumn{1}{c}{GAIC \cite{zeng2019gridanchorbasedimage}} & 
    \multicolumn{1}{c}{Jia \etal \cite{9878802}} & 
    \multicolumn{1}{c}{AesCrop (Ours)} \\

    \includegraphics[width=2cm]{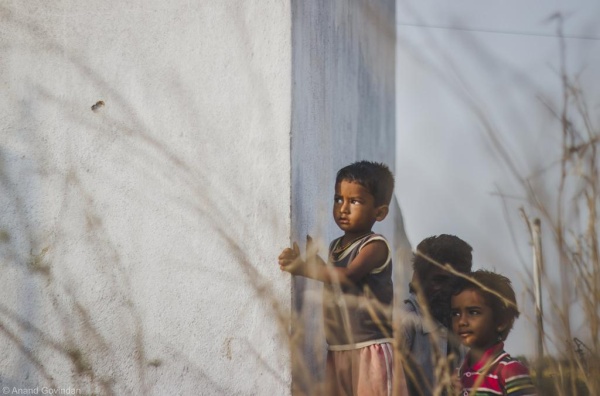} &
    \includegraphics[width=2cm]{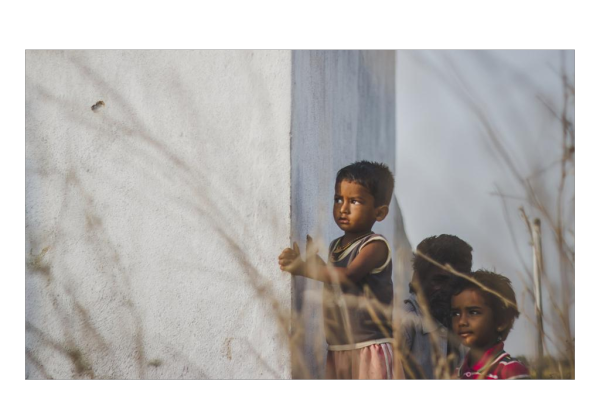} &
    \includegraphics[width=2cm]{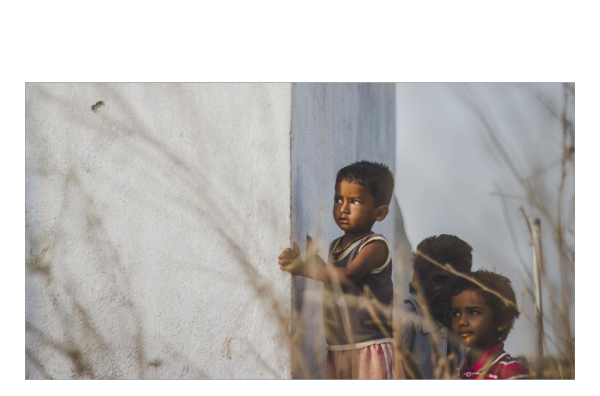} &
    \multicolumn{1}{c|}{-} &
    \includegraphics[width=2cm]{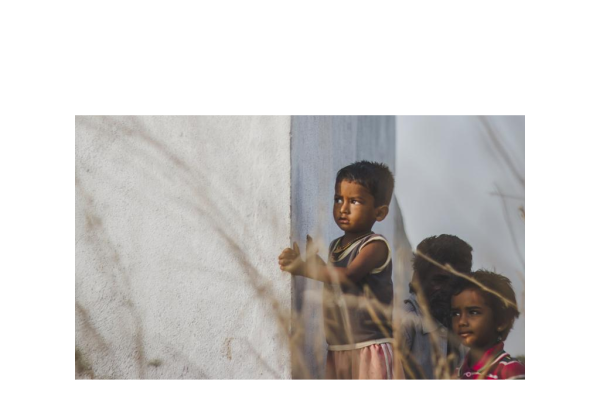} &
    \includegraphics[width=2cm]{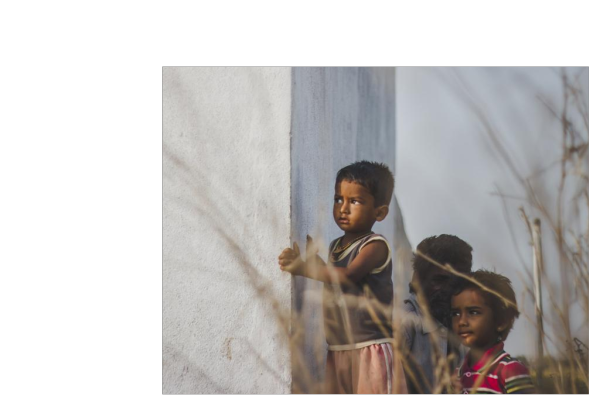} &
    \includegraphics[width=2cm]{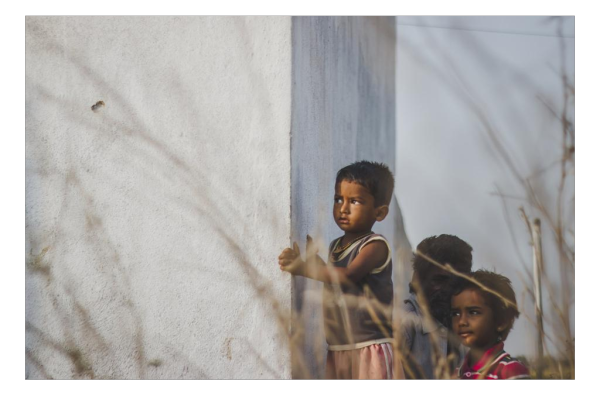} 
    \\ 

    \includegraphics[width=2cm]{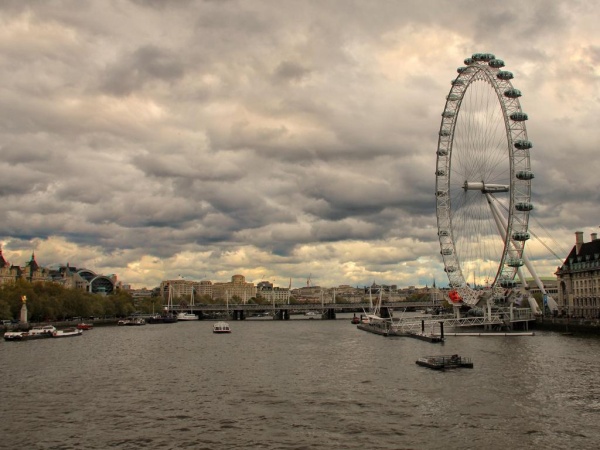} &
    \includegraphics[width=2cm]{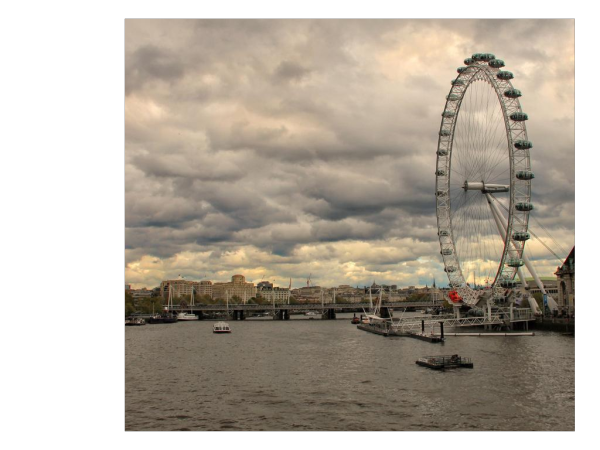} &
    \includegraphics[width=2cm]{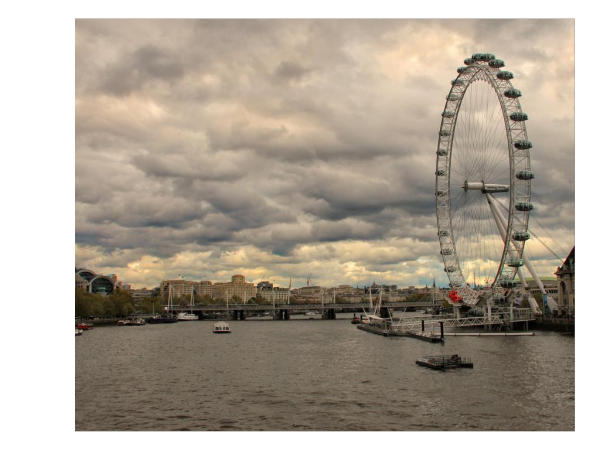} &
    \includegraphics[width=2cm]{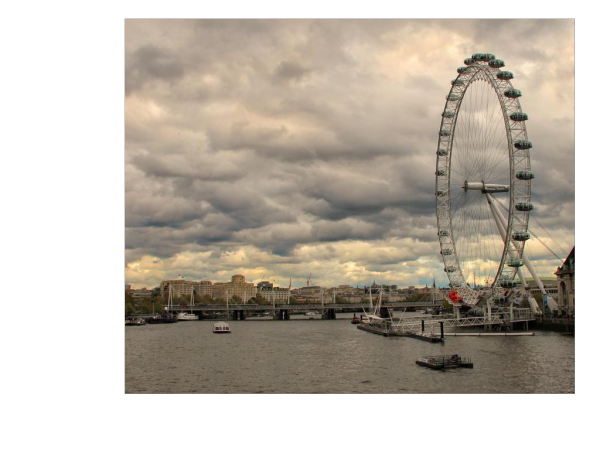} &
    \includegraphics[width=2cm]{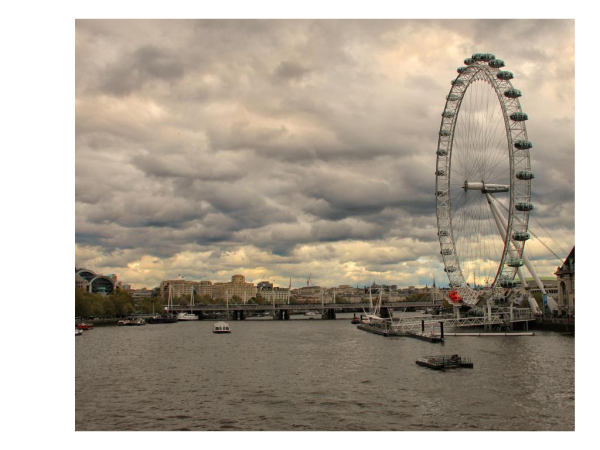} &
    \includegraphics[width=2cm]{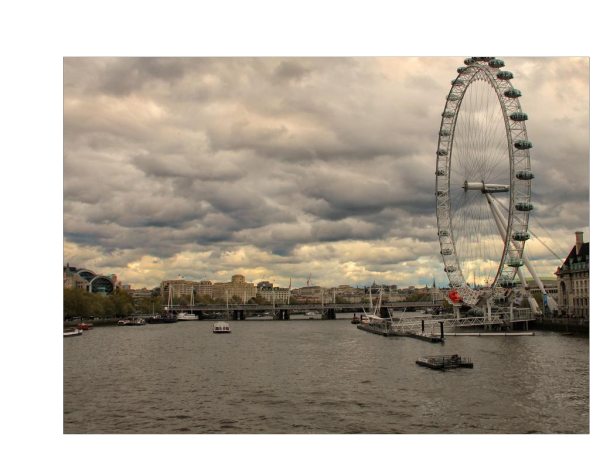} &
    \includegraphics[width=2cm]{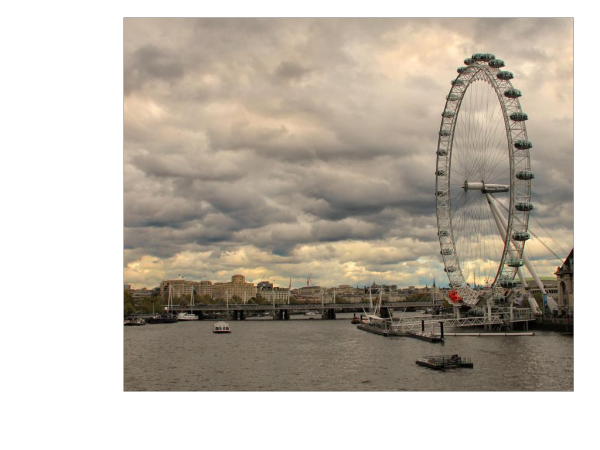} 
    \\ 

    \includegraphics[width=2cm]{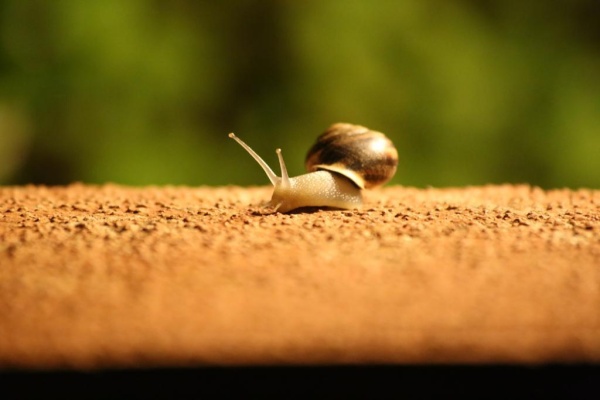} &
    \includegraphics[width=2cm]{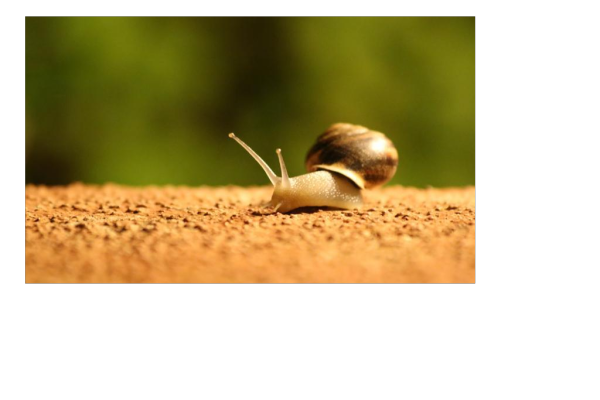} &
    \includegraphics[width=2cm]{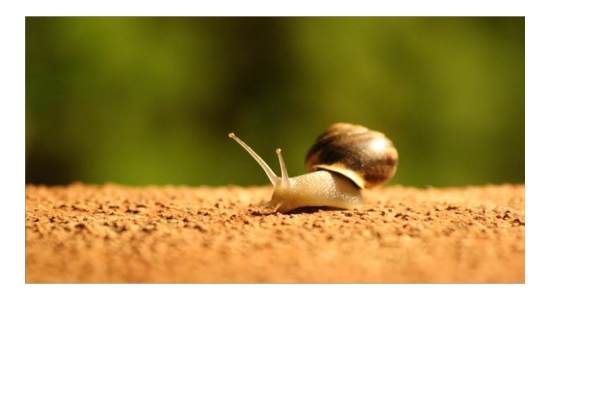} &
    \includegraphics[width=2cm]{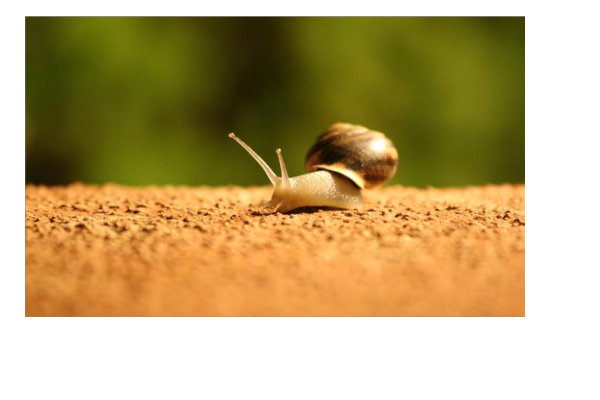} &
    \includegraphics[width=2cm]{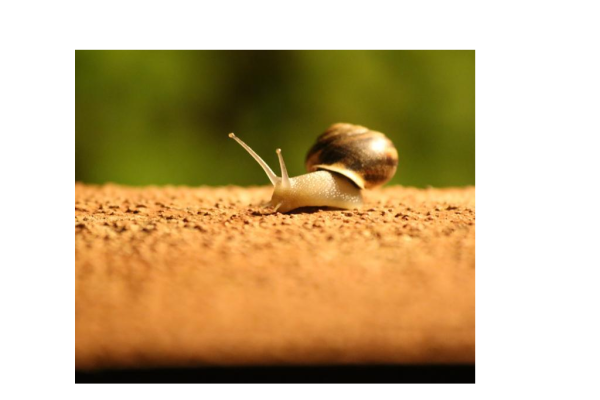} &
    \includegraphics[width=2cm]{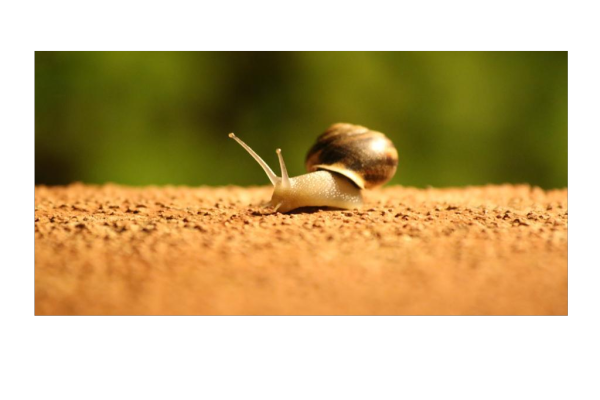} &
    \includegraphics[width=2cm]{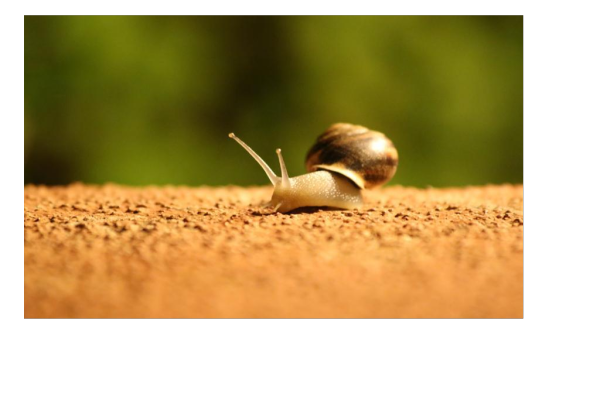} 
    \\ 

    \includegraphics[width=2cm]{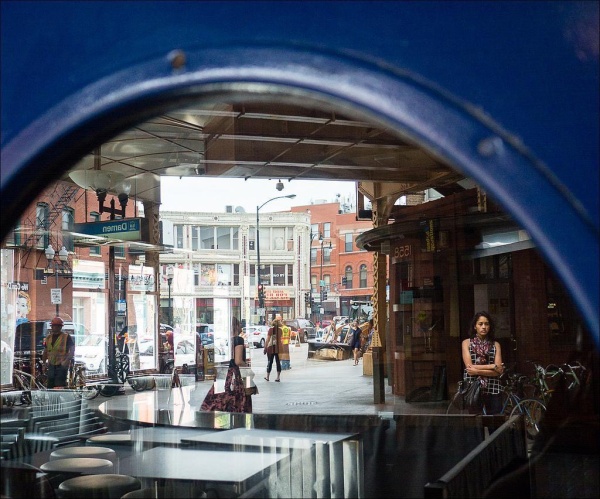} &
    \includegraphics[width=2cm]{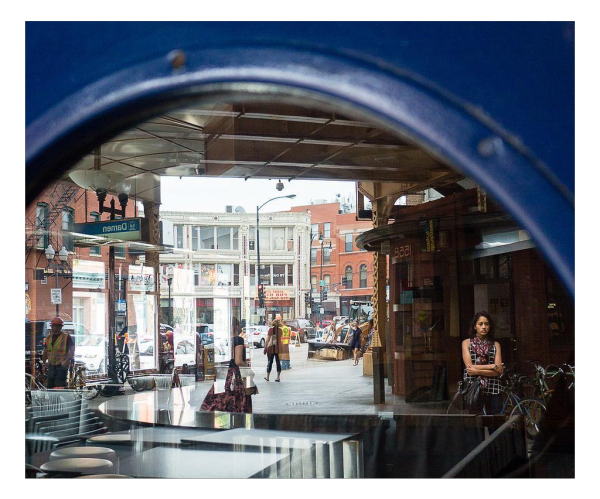} &
    \includegraphics[width=2cm]{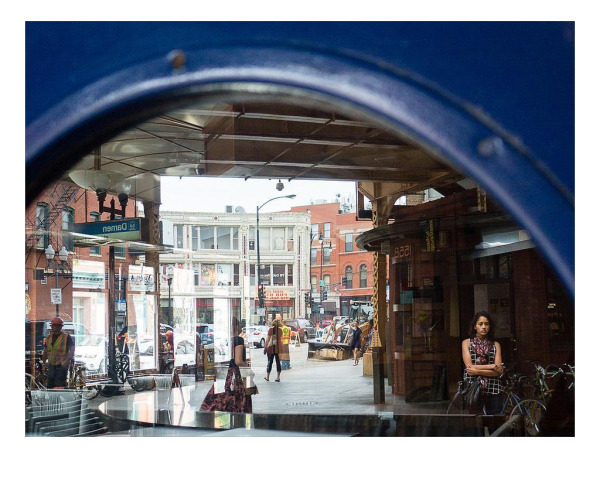} &
    \includegraphics[width=2cm]{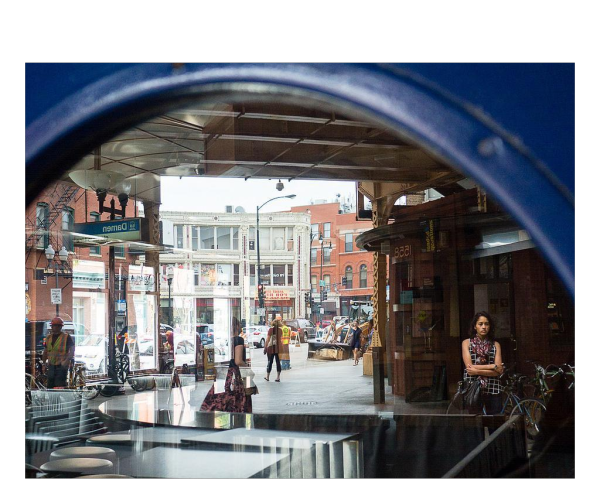} &
    \includegraphics[width=2cm]{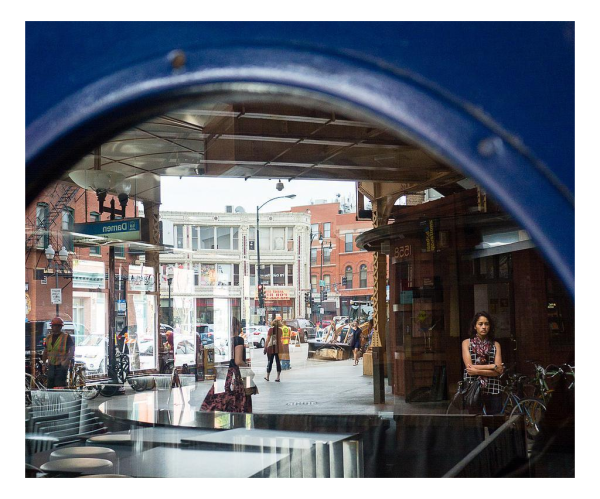} &
    \includegraphics[width=2cm]{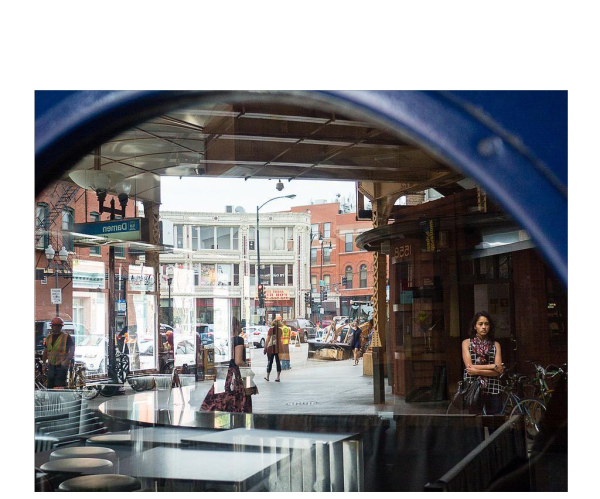} &
    \includegraphics[width=2cm]{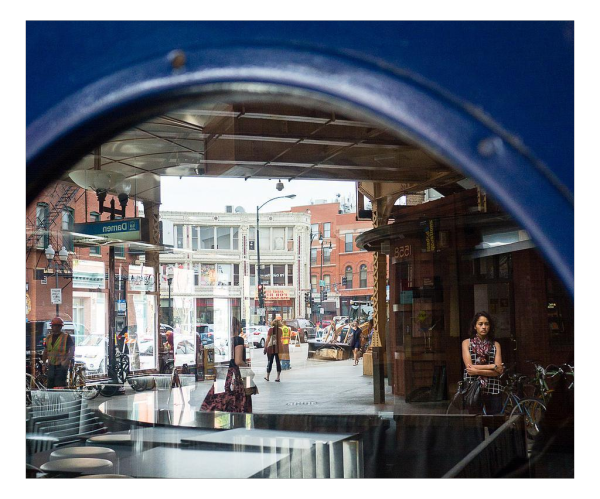} 
    \\ 

    \includegraphics[width=2cm]{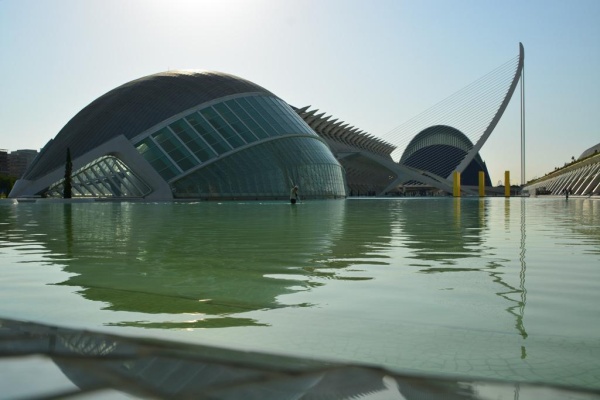} &
    \includegraphics[width=2cm]{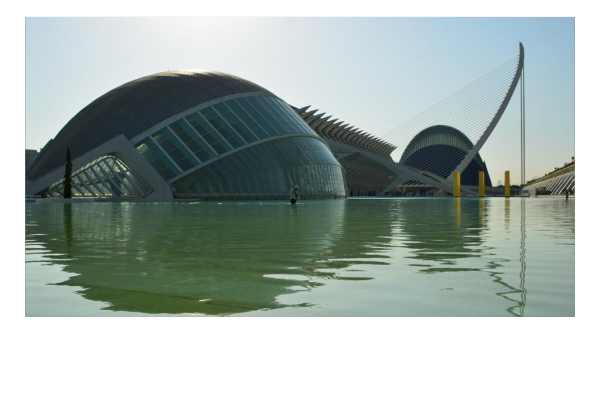} &
    \includegraphics[width=2cm]{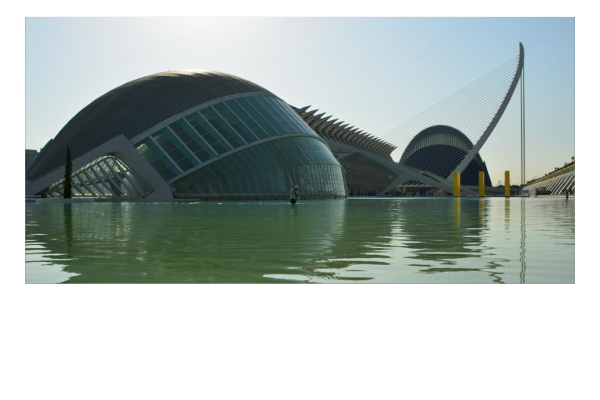} &
    \includegraphics[width=2cm]{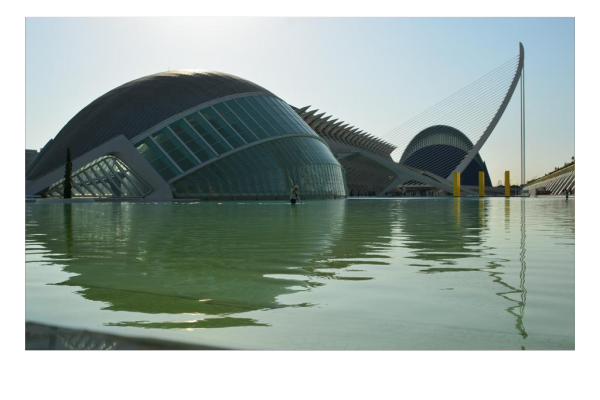} &
    \includegraphics[width=2cm]{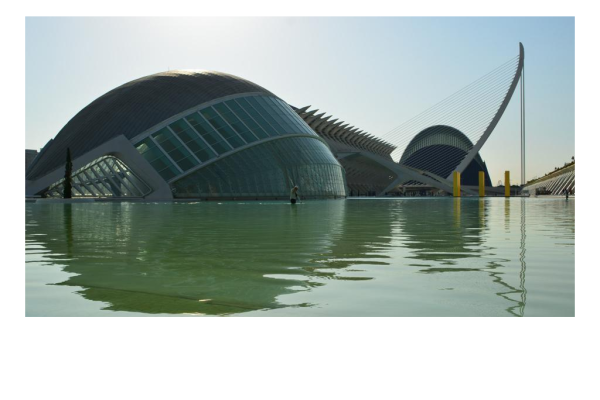} &
    \includegraphics[width=2cm]{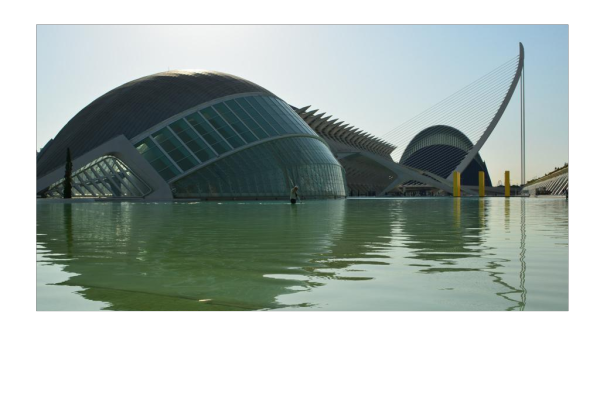} &
    \includegraphics[width=2cm]{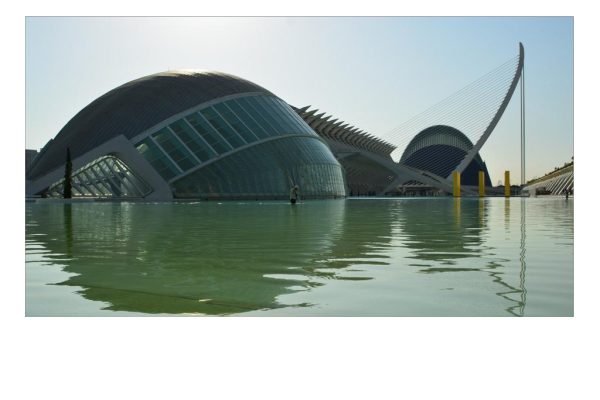} 
    \\ 

    \includegraphics[width=2cm]{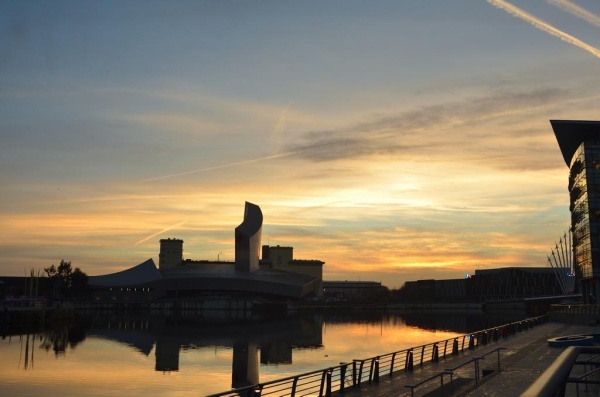} &
    \includegraphics[width=2cm]{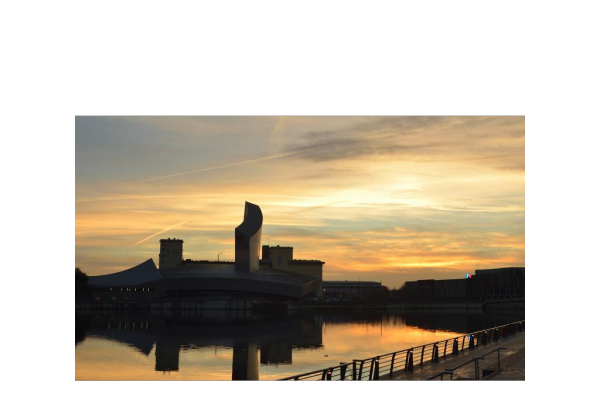} &
    \includegraphics[width=2cm]{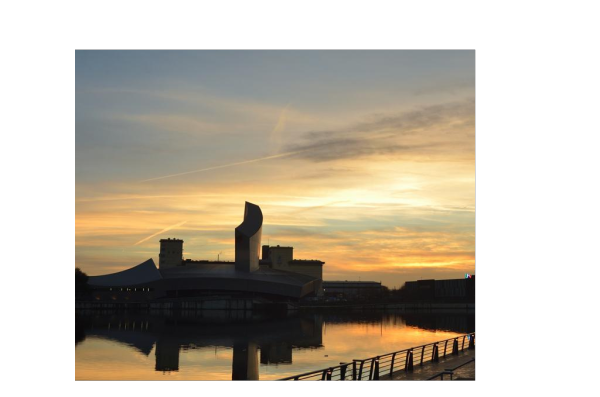} &
    \includegraphics[width=2cm]{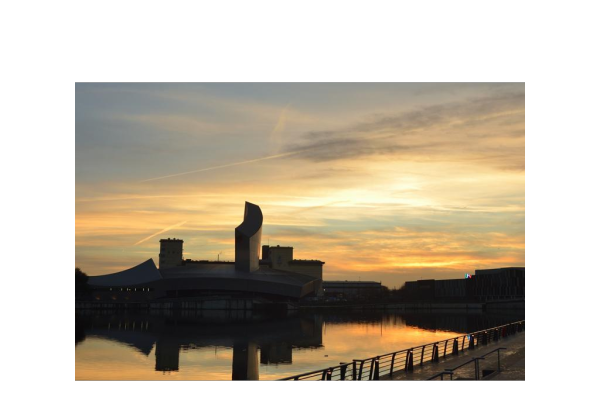} &
    \includegraphics[width=2cm]{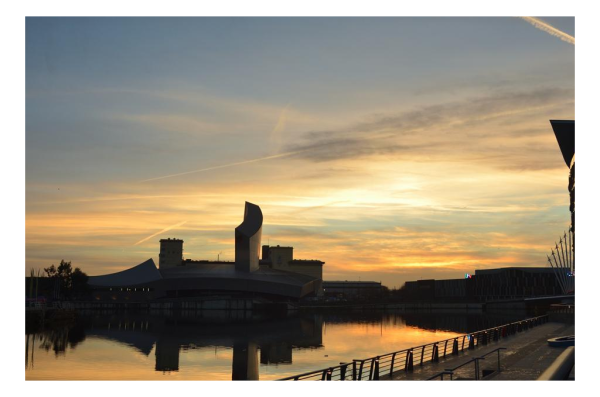} &
    \includegraphics[width=2cm]{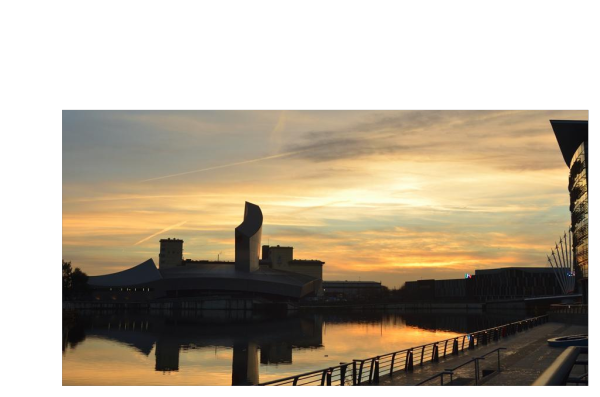} &
    \includegraphics[width=2cm]{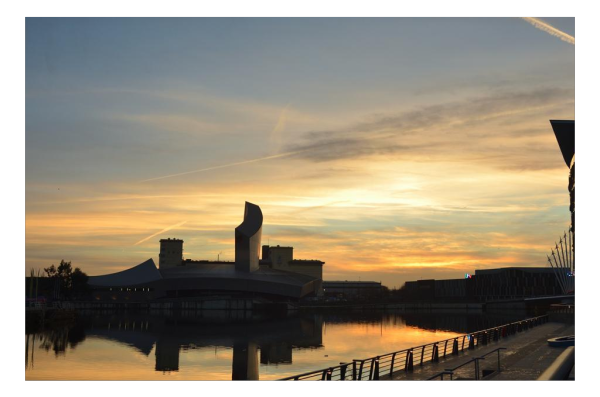} 
    \\ 

    \end{tabular}
    \vspace{-1em}
    \caption{Visual comparison of top three ground truth crops with predicted crops.} 
    \label{fig:qualitative_comp}
\end{figure*}

\indent \textbf{Qualitative Comparison}: Next, we performed a qualitative comparison of AesCrop with two publicly available models, GAIC \cite{zeng2019gridanchorbasedimage} and Jia \etal \cite{9878802}, on six test images from the GAIC \cite{zeng2019gridanchorbasedimage} dataset. For GAIC\cite{zeng2019gridanchorbasedimage}, we generate the crops using the pretrained weights. As Jia \etal \cite{9878802} did not provide pretrained weights, we retrained their model using their original configuration, adjusting the input size to $512 \times 512$ to align with AesCrop for efficient training. For each image, we compare the generated crops with the highest score for each model, alongside the top three ground truth crops with the highest Mean Opinion Scores (MOS), all of which have a score of at least 4. This strategy ensures that we only compare with crops of high quality. \Cref{fig:qualitative_comp} presents a qualitative comparison of these results. 


The visual comparison reveals that AesCrop can generate more aesthetically pleasing  crops compared to the other models in most cases. In the first image, only AesCrop preserved the negative space in front of the children, illustrating a higher correlation with the ground truth crops. Negative space, an often neglected composition rule in automatic cropping approaches, defines the focal point and maintain a sense of balance within the composition. In the second image, AesCrop was also the sole method to remove distracting elements along the left and right borders, effectively directing focus to the Ferris wheel. For the third image, both AesCrop and Jia \etal's \cite{9878802} method successfully eliminated the black border at the bottom, unlike GAIC \cite{zeng2019gridanchorbasedimage}. Comparatively, AesCrop yielded a more aesthetically pleasing result with its placement of the snail adhering to the rule of thirds. In the fourth image, AesCrop and GAIC\cite{zeng2019gridanchorbasedimage} retained the full blue arch which plays a crucial role in framing the scene, while Jia \etal \cite{9878802} cropped it into half, breaking the natrual framing effect. The fifth image showed comparable performance across all models. However, in the sixth image, none of the methods managed to include the rightmost section of the building in their crops. 

\begin{figure}
    \centering
    \begin{tabular}{m{2.4cm} m{2.4cm}} 
    \includegraphics[width=2.4cm]{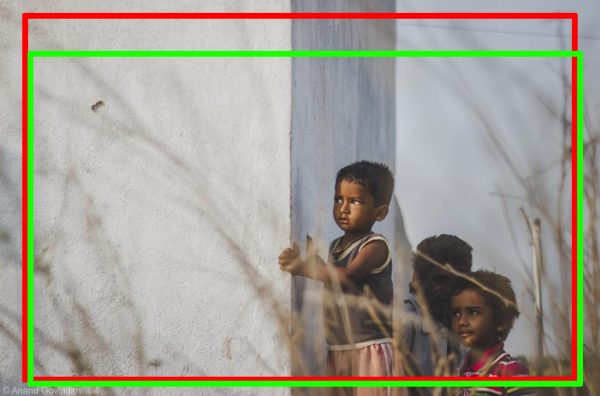} & 
    \includegraphics[width=2.4cm]{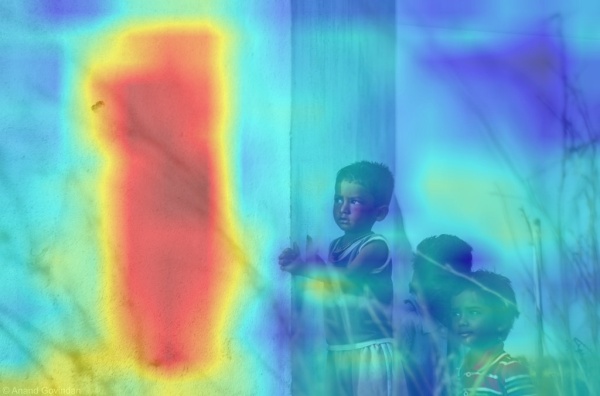} \\
    \includegraphics[width=2.4cm]{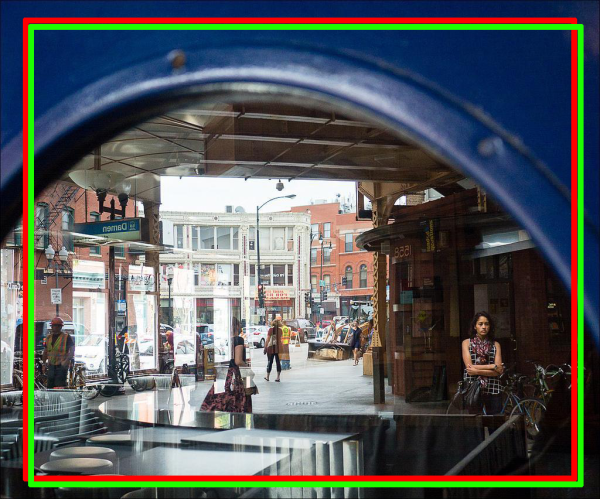} & 
    \includegraphics[width=2.4cm]{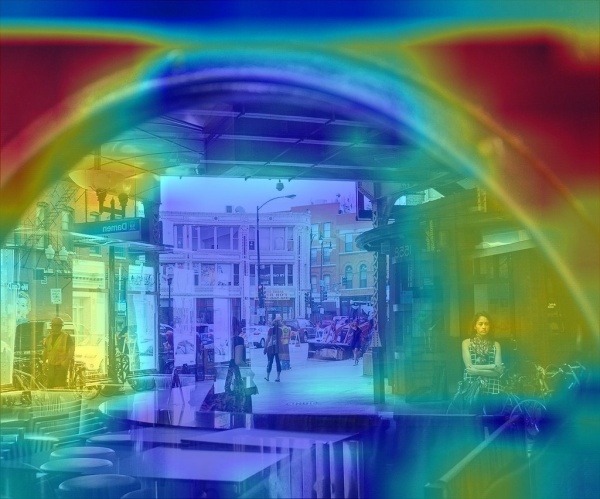} \\
    \end{tabular}
    \vspace{-1em}
    \caption{(left) Source image with ground truth (green box), and AesCrop's best predicted crops (red box) and (right) the corresponding composition attention priors generated by MCAB. } 
    \label{fig:qualitative_mcab}
\end{figure}

We attribute AesCrop's superior performance to the MCAB module. To further analyze the role of MCAB, we selected two examples from \cref{fig:qualitative_comp} and visualize the composition attention bias in \cref{fig:qualitative_mcab}. From the first example, we can observe that MCAB places heavy emphasis on the negative space beside the children, an important compositional region that other models failed to capture. In the second example, MCAB highlights the blue arch, enabling the model to preserve the crucial image framing element. These atention bias heatmaps show that MCAB consistently guides AesCrop to focus on compositionally salient regions, resulting in higher-quality crops.

Overall, AesCrop demonstrates competitive performance, matching or exceeding existing models in most evaluations. It reliably avoids composition failures present in other models, notably reducing over-cropping of key semantic regions and demonstrating a stronger adherence to professional photographic composition rules. We attribute this success to two key innovations: (1) VMamba's ability to generate rich, comprehensive image encodings, and (2) the Mamba Composition Attention Bias (MCAB) mechanism, which effectively directs the model's focus toward compositionally significant areas.

\subsection{Ablation Study}

\hspace{1.2em}\textbf{Importance of Mamba Composition Attention Bias}: The MCAB enhances AesCrop by dynamically modulating attention weights using learned compositional principles. Quantitative analysis in \cref{tab:baseline_comparison} shows that removing MCAB reduces performance by 3.2\% on $Acc_{1/5}$ and 1.4\% on $Acc_{1/10}$, verifying its importance. This performance gap demonstrates that explicit composition modeling through MCAB is crucial for generating aesthetically optimal crops.

\begin{table}
    \centering
    \caption{Impact of MCAB.}
    \vspace{-1em}
    \label{tab:baseline_comparison}
    \begin{tabular}{P{3cm}P{1cm}P{1cm}}
    \hline
    MCAB &
    $Acc_{1/5}$&
    $Acc_{1/10}$ \\ 
    \hline
    \centering \ding{55} & 76.2\% & 90.8\% \\
    \centering \checkmark & 79.4\% & 92.2\% \\ \hline
    \end{tabular}
\end{table}

\textbf{Importance of Object Detection Pretraining}: AesCrop leverages object detection pretraining to boost feature extraction capabilities. As shown in \cref{tab:pretrained_comparison}, this strategy yields significant gains, with the pretrained model outperforms its non-pretrained counterpart by 6.6\% on $Acc_{1/5}$ and 2.6\% on $Acc_{1/10}$. This improvement reflects the pretrained model's superior object localization skills, which enable it to focus on learning aesthetic importance rather than basic visual recognition during cropping-specific training.

\begin{table}
    \centering
    \caption{Impact of Object Detection Pretraining.}
    \vspace{-1em}
    \label{tab:pretrained_comparison}
    \begin{tabular}{P{3cm}P{1cm}P{1cm}}
    \hline
    Pretrained & 
    $Acc_{1/5}$ & 
    $Acc_{1/10}$ \\
    \hline
    \centering \ding{55} & 72.8\% & 89.6\% \\
    \centering \checkmark & 79.4\% & 92.2\% \\ \hline
    \end{tabular}
\end{table}

\textbf{MCAB aggregation technique}: The MCAB attention prior is generated as a probability-weighted average of class activation maps (CAMs). Quantitative comparisons in \cref{tab:cab_type} show that this approach outperforms aggregation approach that considers only the highest-probability CAM, demonstrating that probability-weighted method can better preserve compositional diversity.  On the other hand, the max-probability alternative discards valuable information and introduces noise. This result reinforces the importance of diversity in photographic composition, whereby multiple composition rules can be applied to enhance the aesthetics appeal of a photograph.

\begin{table}
    \centering
    \caption{Impact of MCAB Aggregation Technique.}
    \vspace{-1em}
    \label{tab:cab_type}
    \begin{tabular}{P{4cm}P{1cm}P{1cm}}
    \hline
    Aggregation Technique& 
    $Acc_{1/5}$ & 
    $Acc_{1/10}$ \\ 
    \hline
    \multicolumn{1}{c}{Max} & 76.4\% & 89.0\% \\
    \multicolumn{1}{c}{Average} & 79.4\% & 92.2\% \\ \hline
    \end{tabular}
\end{table}

\begin{figure}
    \centering
    \includegraphics[width=0.35\textwidth]{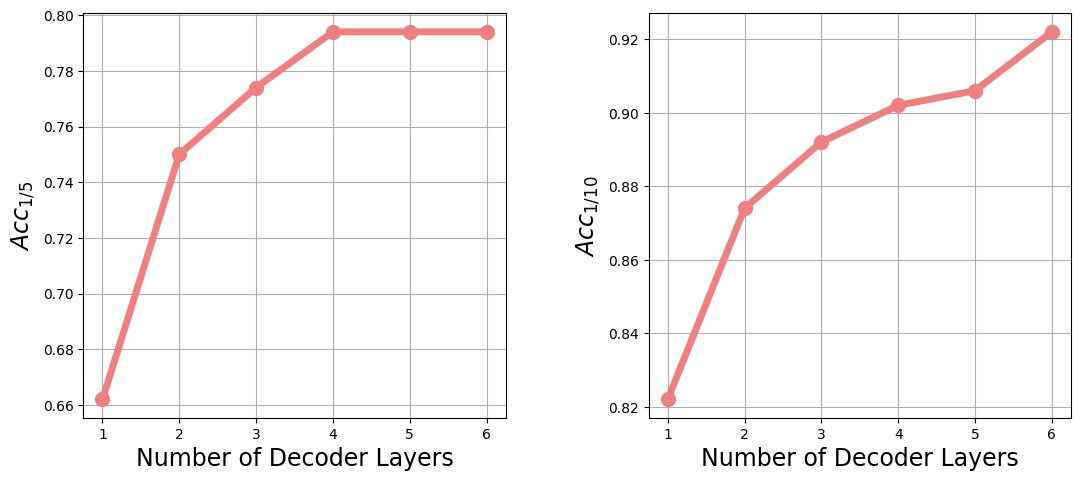}
    \vspace{-1em}
    \caption{Performance across different decoder depths for $Acc_{1/5}$ (left) and $Acc_{1/10}$ (right).} 
    \label{fig:num_decoder}
\end{figure}

\textbf{Number of Decoder Layers}: We empirically determined the optimal number of decoder layers by evaluating configurations from 1 to 6 layers. As \cref{fig:num_decoder} demonstrates, both accuracy metrics peak at 6 layers, achieving 79.4\% for $Acc_{1/5}$ and 92.2\% for $Acc_{1/10}$.

%% file: sec/5_limitations.tex
\section{Limitations}

Although AesCrop demonstrates remarkable performance, it is not without limitations. Since AesCrop was trained on the GAIC \cite{zeng2019gridanchorbasedimage} dataset, where crops are generated by heuristic rules, it inherits the pattern, and thus struggles to achieve true globality.

\begin{figure}
    \centering
    \begin{tabular}{m{3cm} m{3cm}} 
    \includegraphics[width=3cm]{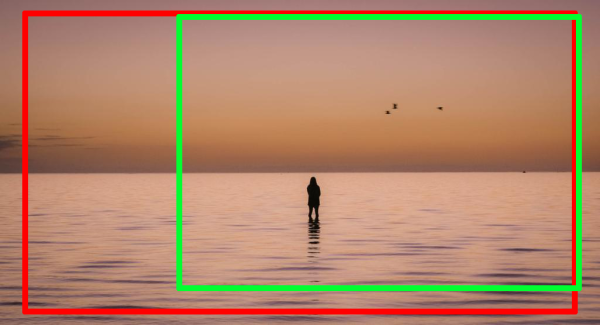} & 
    \includegraphics[width=3cm]{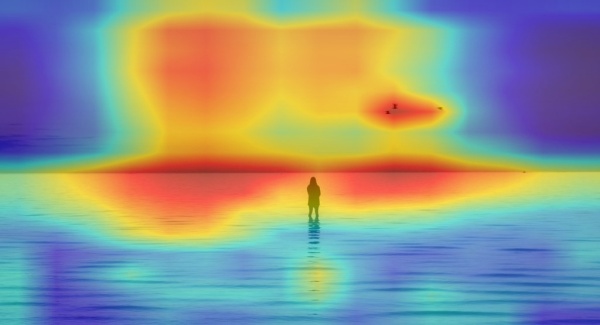} \\
    \end{tabular}
    \vspace{-1em}
    \caption{(left) Source image with ground truth (green box), and AesCrop's best predicted (red box) crops, and (right) AesCrop composition priors generated by MCAB.}
    \label{fig:global_failed}
\end{figure}

Moreover, while the MCAB modulates decoder attention according to the composition structure of the whole image, the optimal crop may not always align with it. For example, \Cref{fig:global_failed} shows a case where the MCAB highlights the central horizontal region, leading AesCrop to prioritize that area, even though the optimal crop does not rely on this particular compositional feature.

%% file: sec/6_conclusion.tex
\section{Conclusion}

We introduce AesCrop, a hybrid cropping framework that incorporates compositional knowledge via Mamba Composition Attention Bias. By directing decoder attention toward compositionally salient regions, AesCrop achieves superior alignment with human annotations across all metrics, achieving state-of-the-arts performance. Extensive experiments validate the impact of each architectural component while revealing two key limitations: heuristic-induced biases from training data and occasional over-prioritization of compositional cues. These insights suggest promising directions for future work to enhance global adaptability and further refine aesthetic and compositional integration.

%% file: main.bbl
\begin{thebibliography}{19}
\providecommand{\natexlab}[1]{#1}
\providecommand{\url}[1]{\texttt{#1}}
\expandafter\ifx\csname urlstyle\endcsname\relax
  \providecommand{\doi}[1]{doi: #1}\else
  \providecommand{\doi}{doi: \begingroup \urlstyle{rm}\Url}\fi

\bibitem[Chen et~al.(2017)Chen, Klopp, Sun, Chien, and Ma]{chen-acmmm-2017}
Yi-Ling Chen, Jan Klopp, Min Sun, Shao-Yi Chien, and Kwan-Liu Ma.
\newblock Learning to compose with professional photographs on the web.
\newblock In \emph{ACM Multimedia 2017}, 2017.

\bibitem[Guo et~al.(2018)Guo, Wang, Shen, Yan, and Liao]{8259308}
Guanjun Guo, Hanzi Wang, Chunhua Shen, Yan Yan, and Hong-Yuan~Mark Liao.
\newblock Automatic image cropping for visual aesthetic enhancement using deep
  neural networks and cascaded regression.
\newblock \emph{IEEE Transactions on Multimedia}, 20\penalty0 (8):\penalty0
  2073--2085, 2018.

\bibitem[Hong et~al.(2021)Hong, Du, Xian, Lu, Cao, and Zhong]{9578088}
Chaoyi Hong, Shuaiyuan Du, Ke Xian, Hao Lu, Zhiguo Cao, and Weicai Zhong.
\newblock Composing photos like a photographer.
\newblock In \emph{2021 IEEE/CVF Conference on Computer Vision and Pattern
  Recognition (CVPR)}, pages 7053--7062, 2021.

\bibitem[Hong et~al.(2024)Hong, Yuan, Gharbi, Fisher, and
  Fatahalian]{hong2024learningsubjectawarecroppingoutpainting}
James Hong, Lu Yuan, Michaël Gharbi, Matthew Fisher, and Kayvon Fatahalian.
\newblock Learning subject-aware cropping by outpainting professional photos,
  2024.

\bibitem[Jia et~al.(2022)Jia, Huang, Fu, and He]{9878802}
Gengyun Jia, Huaibo Huang, Chaoyou Fu, and Ran He.
\newblock Rethinking image cropping: Exploring diverse compositions from global
  views.
\newblock In \emph{2022 IEEE/CVF Conference on Computer Vision and Pattern
  Recognition (CVPR)}, pages 2436--2445, 2022.

\bibitem[Lee et~al.(2018)Lee, Kim, Lee, and Kim]{LEE201891}
Jun-Tae Lee, Han-Ul Kim, Chul Lee, and Chang-Su Kim.
\newblock Photographic composition classification and dominant geometric
  element detection for outdoor scenes.
\newblock \emph{Journal of Visual Communication and Image Representation},
  55:\penalty0 91--105, 2018.

\bibitem[Li et~al.(2020)Li, Zhang, and Huang]{9156674}
Debang Li, Junge Zhang, and Kaiqi Huang.
\newblock Learning to learn cropping models for different aspect ratio
  requirements.
\newblock In \emph{2020 IEEE/CVF Conference on Computer Vision and Pattern
  Recognition (CVPR)}, pages 12682--12691, 2020.

\bibitem[Liu et~al.(2024)Liu, Tian, Zhao, Yu, Xie, Wang, Ye, Jiao, and
  Liu]{liu2024vmambavisualstatespace}
Yue Liu, Yunjie Tian, Yuzhong Zhao, Hongtian Yu, Lingxi Xie, Yaowei Wang,
  Qixiang Ye, Jianbin Jiao, and Yunfan Liu.
\newblock Vmamba: Visual state space model, 2024.

\bibitem[Meng et~al.(2021)Meng, Chen, Fan, Zeng, Li, Yuan, Sun, and
  Wang]{DBLP:journals/corr/abs-2108-06152}
Depu Meng, Xiaokang Chen, Zejia Fan, Gang Zeng, Houqiang Li, Yuhui Yuan, Lei
  Sun, and Jingdong Wang.
\newblock Conditional {DETR} for fast training convergence.
\newblock \emph{CoRR}, abs/2108.06152, 2021.

\bibitem[Pan et~al.(2021)Pan, Cao, Wang, Lu, and Zhong]{9710744}
Zhiyu Pan, Zhiguo Cao, Kewei Wang, Hao Lu, and Weicai Zhong.
\newblock Transview: Inside, outside, and across the cropping view boundaries.
\newblock In \emph{2021 IEEE/CVF International Conference on Computer Vision
  (ICCV)}, pages 4198--4207, 2021.

\bibitem[Pan et~al.(2023)Pan, Chen, Zhang, Lu, Cao, and
  Zhong]{pan2023beautyrarecontrastivecomposition}
Zhiyu Pan, Yinpeng Chen, Jiale Zhang, Hao Lu, Zhiguo Cao, and Weicai Zhong.
\newblock Find beauty in the rare: Contrastive composition feature clustering
  for nontrivial cropping box regression, 2023.

\bibitem[Samet et~al.(2020)Samet, Hicsonmez, and Akbas]{HoughNet}
Nermin Samet, Samet Hicsonmez, and Emre Akbas.
\newblock Houghnet: Integrating near and long-range evidence for bottom-up
  object detection.
\newblock In \emph{European Conference on Computer Vision (ECCV)}, 2020.

\bibitem[Selvaraju et~al.(2016)Selvaraju, Das, Vedantam, Cogswell, Parikh, and
  Batra]{DBLP:journals/corr/SelvarajuDVCPB16}
Ramprasaath~R. Selvaraju, Abhishek Das, Ramakrishna Vedantam, Michael Cogswell,
  Devi Parikh, and Dhruv Batra.
\newblock Grad-cam: Why did you say that? visual explanations from deep
  networks via gradient-based localization.
\newblock \emph{CoRR}, abs/1610.02391, 2016.

\bibitem[Shi et~al.(2023)Shi, Chen, He, Song, and Hao]{10222223}
Tengfei Shi, Chenglizhao Chen, Yuanbo He, Wenfeng Song, and Aimin Hao.
\newblock Joint probability distribution regression for image cropping.
\newblock In \emph{2023 IEEE International Conference on Image Processing
  (ICIP)}, pages 990--994, 2023.

\bibitem[Wei et~al.(2018)Wei, Zhang, Shen, Lin, Mech, Hoai, and
  Samaras]{wei2018good}
Zijun Wei, Jianming Zhang, Xiaohui Shen, Zhe Lin, Radomir Mech, Minh Hoai, and
  Dimitris Samaras.
\newblock Good view hunting: Learning photo composition from dense view pairs.
\newblock In \emph{Proceedings of IEEE Conference on Computer Vision and
  Pattern Recognition}, 2018.

\bibitem[Yan et~al.(2013)Yan, Lin, Kang, and Tang]{6618974}
Jianzhou Yan, Stephen Lin, Sing~Bing Kang, and Xiaoou Tang.
\newblock Learning the change for automatic image cropping.
\newblock In \emph{2013 IEEE Conference on Computer Vision and Pattern
  Recognition}, pages 971--978, 2013.

\bibitem[Yuan et~al.(2024)Yuan, Li, and Chen]{YUAN2024104316}
Quan Yuan, Leida Li, and Pengfei Chen.
\newblock Aesthetic image cropping meets vlp: Enhancing good while reducing
  bad.
\newblock \emph{Journal of Visual Communication and Image Representation},
  105:\penalty0 104316, 2024.

\bibitem[Zeng et~al.(2019)Zeng, Li, Cao, and
  Zhang]{zeng2019gridanchorbasedimage}
Hui Zeng, Lida Li, Zisheng Cao, and Lei Zhang.
\newblock Grid anchor based image cropping: A new benchmark and an efficient
  model, 2019.

\bibitem[Zhong et~al.(2023)Zhong, Cheng, Wu, Yuan, Zheng, Li, Hu, Lin, Sato,
  and Sato]{10350920}
Zhihang Zhong, Mingxi Cheng, Zhirong Wu, Yuhui Yuan, Yinqiang Zheng, Ji Li, Han
  Hu, Stephen Lin, Yoichi Sato, and Imari Sato.
\newblock Clipcrop: Conditioned cropping driven by vision-language model.
\newblock In \emph{2023 IEEE/CVF International Conference on Computer Vision
  Workshops (ICCVW)}, pages 294--304, 2023.

\end{thebibliography}
